\documentclass[letterpaper, 10 pt, conference]{ieeeconf}  

\IEEEoverridecommandlockouts                              

\let\labelindent\relax
\usepackage{graphicx}
\usepackage{amssymb,mathrsfs,amsmath}
\usepackage{pifont}
\usepackage{enumitem}
\usepackage{makecell}
\usepackage{threeparttable}
\usepackage{booktabs}
\usepackage{multirow}
\usepackage{color}
\usepackage[switch]{lineno}

\newcommand{\red}[1]{\textcolor{black}{#1}}
\newcommand{\blue}[1]{\textcolor{black}{#1}}

\title{\LARGE \bf
Perception Imitation: \\Towards \red{Synthesis-free} Simulator for Autonomous Vehicles
}

\author{Xiaoliang Ju$^{1,2}$, Yiyang Sun$^{2,3}$,  Yiming Hao$^{4}$, Yikang Li$^{2}$, Yu Qiao$^{2}$, and Hongsheng Li$^{1,\dag}$
\thanks{$^{1}$ X. Ju and H. Li are with the Multimedia Laboratory, The Chinese University of Hong Kong, Hong Kong.}%
\thanks{$^{2}$ X. Ju, Y. Sun, Y. Li, and Y. Qiao are with the Shanghai Artificial Intelligence Laboratory, Shanghai, China.}%
\thanks{$^{3}$ Y. Sun is also with the Tongji University, Shanghai, China.}%
\thanks{$^{4}$ Y. Hao is with the Centre for Perceptual and Interactive Intelligence Limited, Hong Kong.}
\thanks{$^{\dag}$ The corresponding author.}
}

\begin{document}

\maketitle
\thispagestyle{empty}
\pagestyle{empty}

\begin{abstract}
    We propose a perception imitation method to simulate results of a certain perception model, and discuss a new heuristic route of autonomous driving simulator without data synthesis. The motivation is that original sensor data is not always necessary for tasks such as planning and control when semantic perception results are ready, so that simulating perception directly is more economic and efficient. In this work, a series of evaluation methods such as matching metric and performance of downstream task are exploited to examine the simulation quality. Experiments show that our method is effective to model the behavior of learning-based perception model, and can be further applied in the proposed simulation route smoothly. 
    
\end{abstract}



\section{Introduction}
	
    Autonomous driving has been one of the most popular applications in robotics, with lots of related products landed in industrial market~\cite{kirkpatrick2022still}. To make the system robust and safe, repetitious tests in combination with various cases need to be conducted regularly, where driving simulator plays an important role for problem pre-filtering and solution validation before real-car test.
    
\begin{figure}[!t]
        \centering
        \includegraphics[trim={2.1cm 4.5cm 3cm 4cm},clip, width=8.5cm]{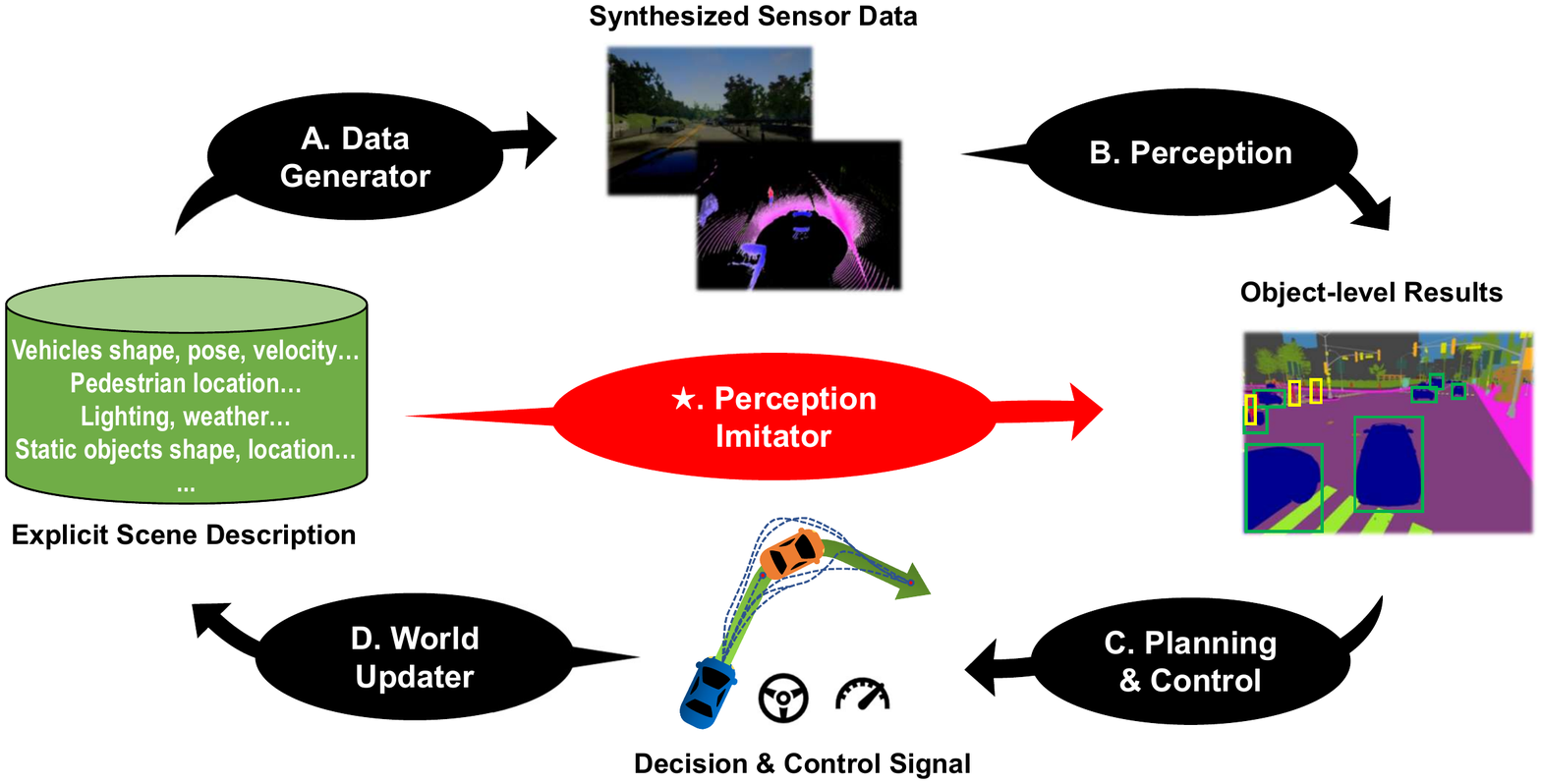}
        \caption{\red{Simulator loop from the perspective of ego-vehicle. $A\rightarrow B\rightarrow C\rightarrow D\rightarrow A$ is the traditional simulation loop, whereas $\bigstar \rightarrow C \rightarrow D \rightarrow \bigstar$ denotes the new loop equipped with our perception imitator.}
}
        \label{fig:pipeline}
    \end{figure}

    Classical simulators such as CarSim~\cite{kinjawadekar2009vehicle} and SUMO~\cite{lopez2018microscopic} are favored by traditional carmakers, but their objectives are often incompatible with the new trend of autonomous driving. For example, CarSim~\cite{kinjawadekar2009vehicle} is good at simulating vehicle dynamics, and SUMO~\cite{lopez2018microscopic} is mainly used for traffic network modeling. 
    There are also game-like simulators which care more about control, such as racing simulator TORCS~\cite{wymann2000torcs} and RL~(reinforcement learning) oriented MetaDrive~\cite{li2021metadrive}, but their traffic scenarios are too simple.
    \red{
   Compared to these simulators that only focus on special parts, CARLA~\cite{dosovitskiy2017carla} becomes popular for covering a full pipeline of autonomous driving. 
   As the outer loop shown in Figure~\ref{fig:pipeline}, the simulation of ego vehicle is performed as a cycle of data generation, perception, control, and world update. 
    The main problems of CARLA-like simulators are two folds. 
    First, such a full pipeline leads to heavy system workload, especially the data generation process.
    More importantly, unrealistic data synthesis may bring considerable domain gap to the downstream modules.
    }
    
\begin{figure*}[!h]
    \centering
        \includegraphics[trim={2.6cm 9.2cm 3.2cm 7.2cm},clip, width=17.5cm]{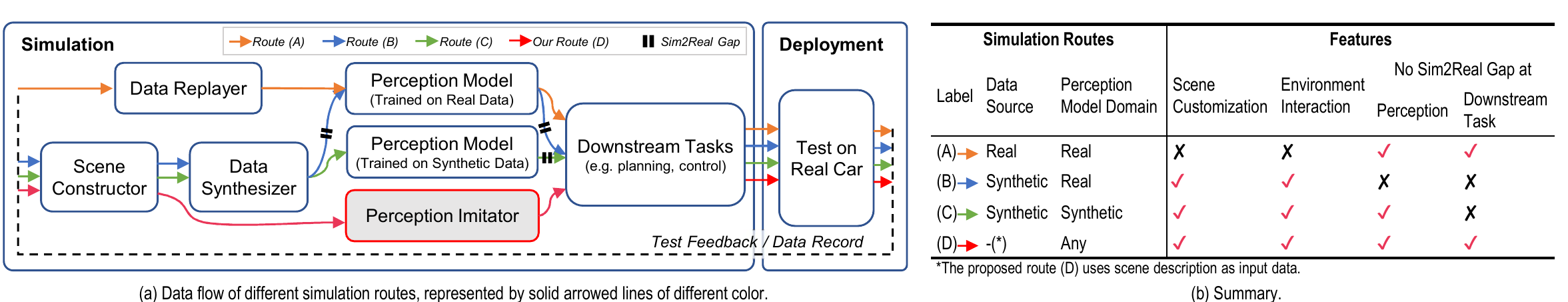}
    \caption{\red{Comparison of Different Simulation Routes.}}
    \label{fig:route}
    \vspace{-4mm}
    \end{figure*}

    \red{
    To further clarify the essential problems in driving simulation, several typical simulation routes are enumerated as the Figure \ref{fig:route} (a) shows. Route (A) is the most basic one for case debugging by replaying real data record without any Sim2Real (simulation to reality) gap, but the data is not editable once recorded, so that no new agent-environment interactions can be produced. The interaction means the environment should give response to any action performed by the ego vehicle, and vice versa. 
    To enable the interactions, scene constructor and data synthesizer are needed to generate new data according to the results of frame-by-frame interactions as Route (B). However, once synthetic data is involved, a Sim2Real gap between the data source and the perception model emerges. An alternative is to train the perception model on synthetic data as Route (C), but the gap will still emerge at the downstream tasks as the perception domain changes. Although techniques such as DA~(domain adaptation)~\cite{bewley2019learning} and DR~(domain randomization)~\cite{loquercio2019deep} are proposed to bridge those gaps, they often serve as complementary remedies without resolving the gap from source.
    }

    \red{Obviously, the source of those Sim2Real problems is data reality, which is always a hard nut to crack. However, the sensor data is not necessary for some tasks such as planning and control. Therefore, we conceive a novel route which helps to release the burden of data synthesis, while avoiding the Sim2Real problem. 
    The key idea is to build a map from traffic scene description to the output of a target perception model, so that the data reality problem can be avoided.
    We call this mapping module as {\bf perception imitator}. 
    Equipped with a perception imitator, the CARLA-like simulation loop can be simplified as Figure \ref{fig:pipeline} shows, with the corresponding data flow as Route (D) in Figure \ref{fig:route}.
    }

    More specifically, our contributions lie in the following aspects:
    \begin{itemize}
        \item A heuristic simulation route is designed for tasks that do not rely on raw data, which exploits a perception imitator to predict perception results of a specific~model.
        \item A scene description is devised as input for perception imitator to explore its relevance to perception results.
        \item We develop a CNN-based perception imitator model to predict results of popular 3D detection models. 
        \item We define a series of metrics to evaluate our perception imitator, such as matching quality on different datasets and the performance of downstream task.  Experiments show that our model is efficient to imitate the target perception model on all these metrics.
    \end{itemize}

\section{Related Works}
\label{sec:review}
\subsection{Perception \& Planning}

\red{In our work, we care more about how to imitate a target perception model for simulation Route(D) in Figure \ref{fig:route}, rather than the original performance of perception, planning and control. Therefore, we only give a brief review in these areas due to limited space.}

Plethora of works investigate the perception pipeline leveraging various sensors like camera, LiDAR and radar~\cite{nabati2021centerfusion}\cite{wang2022detr3d}. As one of the mainstream routes, deep learning based 3D detection has shown good performance and become dominant in recent years~\cite{lang2019pointpillars}\cite{shi2020pv}\cite{yin2021center}.
As such, the object-level perception results can be directly transformed to a representation of occupancy and input to the planning module.

 Motion planning has long been treated as an optimization problem defined on manual cost functions, such as \cite{ajanovic2018search} adopts A* algorithm for path searching, and \cite{fan2018baidu} exploits dynamic programming to optimize the path and speed control. With the advances in reinforcement learning and imitation learning, neural networks are employed as action controller~\cite{kendall2019learning}\cite{chen2019deep}. As the traditional methods suffer from massive engineering fine-tunes for various cases, the learning-based ones are more promising to deal with complex scenarios in a more direct~way.

\subsection{Uncertainty and Error Modeling} An intuitive idea to predict the output of deep neural network is to model the uncertainty or error distribution of its output~\cite{gal2016dropout}\cite{lakshminarayanan2017simple}\cite{guo2017calibration}. \cite{gal2016dropout} proposes a method using multiple inference with dropout to get an output distribution sampling.~\cite{guo2017calibration} uses calibration method to model the result deviation from the ground truth. \blue{The problem is most of these works are limited to low-dimension tasks such as classification. Most other approaches of detection uncertainty modeling are often coupled with their own network structure~\cite{lu2021geometry}\cite{feng2021review}, and thus hard to be transferred to other models. Moreover, these methods need original sensor data as input, which precludes it to be applied in our tasks.}

\subsection{Knowledge Distillation} This technique is often used in model compression~\cite{hinton2015distilling}. The idea is to transfer knowledge from a complex teacher model to a simple student model, expecting the student to approach or even surpass the teacher. Plenty of works have achieved great results in recent years~\cite{chen2019data}\cite{yin2020dreaming}. \blue{Although it is similar to our task in the need to learn from the target model, there are two key differences. First, different types of knowledge such as feature, network parameter can be exploited to improve the student model~\cite{gou2021knowledge}, while target model in our tasks should be a black box.  Second, the student model in knowledge distillation takes the same training data as input, while we want to predict the target model behavior without any original sensor data.}

\red{\subsection{Perception Imitation}
\label{sub:pi}
To the best of our knowledge, only \cite{wong2020testing} meet our requirements mentioned before. The authors propose a perception simulation method to test the safety of planning modules, with similar input and output definitions as ours. As such, it is selected as our baseline.} 

\red{
Different from \cite{wong2020testing}, we aim to propose a new simulation route without data synthesis, which is realized by employing a perception imitator.  Second, our imitator model exploits different scene description unit and training loss to achieve an accuracy leap on matching metrics. Experiments show that the baseline model is too weak to explore knowledge embedded in the scene description. Third, all perception models and datasets used in our evaluation are open source and popular in recent years.  In addition, we involve interactive online tasks rather than only offline ones to demonstrate the application potential in the simulator.
}



\section{\red{Synthesis-free} Simulation}
Before formulating our novel \red{synthesis-free} simulation route, we will first revisit the traditional route depending on data synthesis, as our basis.
Here we only discuss the self-driving simulation from the perspective of the ego-vehicle. 

\subsection{Revisiting the Traditional Simulation}
A classical simulation loop can be roughly divided into the following 4 steps as Figure.\ref{fig:pipeline} shows. 
    \subsubsection{Data Generation} At the time-step $t$, once the \red{ego vehicle} is set up to pose $x_t$, the sensor data $d_t$ can be generated from the scene description $s_t$ by a projection function $h$.
    \begin{equation}
    d_t = h(s_t, x_t)
    \end{equation}
    \subsubsection{Perception}   Perception model $f_{\Theta}$ with parameter set $\Theta$ such as 3D detection is used to extract object information from sensor data $d_t$. The perception result is denoted as $c_t$.
    \begin{equation}
    c_t = f_{\Theta}(d_t)
    \end{equation}
    \subsubsection{Planning \& Control} With the object-level perception results $c_t$, planning and control module $q$ makes action $a_t$.
    \begin{equation}
        a_t = q(c_t)
    \end{equation}
    \subsubsection{Scene Update}  When action $a_t$ is done, the scene description $s_t$ can be updated to $s_{t+1}$ by \red{the transition function ~$\pi$ in simulator}. 
    \begin{equation}
        s_{t+1} = \pi(s_t, a_t)
    \end{equation}

\subsection{Discarding Data Dependency by Perception Imitator}
Here we discuss how to build the \red{synthesis-free} simulation based on the traditional framework.
In contrast to the traditional loop, we use a perception imitator ($\bigstar$ in Figure \ref{fig:pipeline}) to substitute the first 2 steps of the traditional loop. The new loop can be summarized as
\begin{equation}
\begin{aligned}
    \hat{c_t} &= \hat{f}(s_t; f_{\Theta})\\
    a_t & = q(\hat{c_t})\\
    s_{t+1} &= \pi(s_t, a_t)
\end{aligned}
\end{equation}
where $\hat{c_t}$ is the simulated perception results, $\hat{f}$ is our perception imitator. 

In our formulation, the input of $\hat{f}$ is the scene state description $s_t$, and the output is simulated perception results $\hat{c_t}$, such as a series of bounding box coordinates.
The imitation target of $\hat{f}$ is the specific perception model $f_\Theta$, so that we use $\hat{f}(s_t; f_{\Theta})$ to represent that the parameters of $\hat{f}$ is conditioned on $f_\Theta$. \red{Our perception imitator $\hat{f}$ actually is a mapping from the scene description to the behavior of perception model, the feasibility of which is discussed at the end of subsection \ref{subsec:scene description}.}


\section{Perception Imitator}
\begin{figure}[!t]
    \centering
    \includegraphics[trim={1.4cm 10.5cm 10.8cm 3.8cm},clip, width=\linewidth]{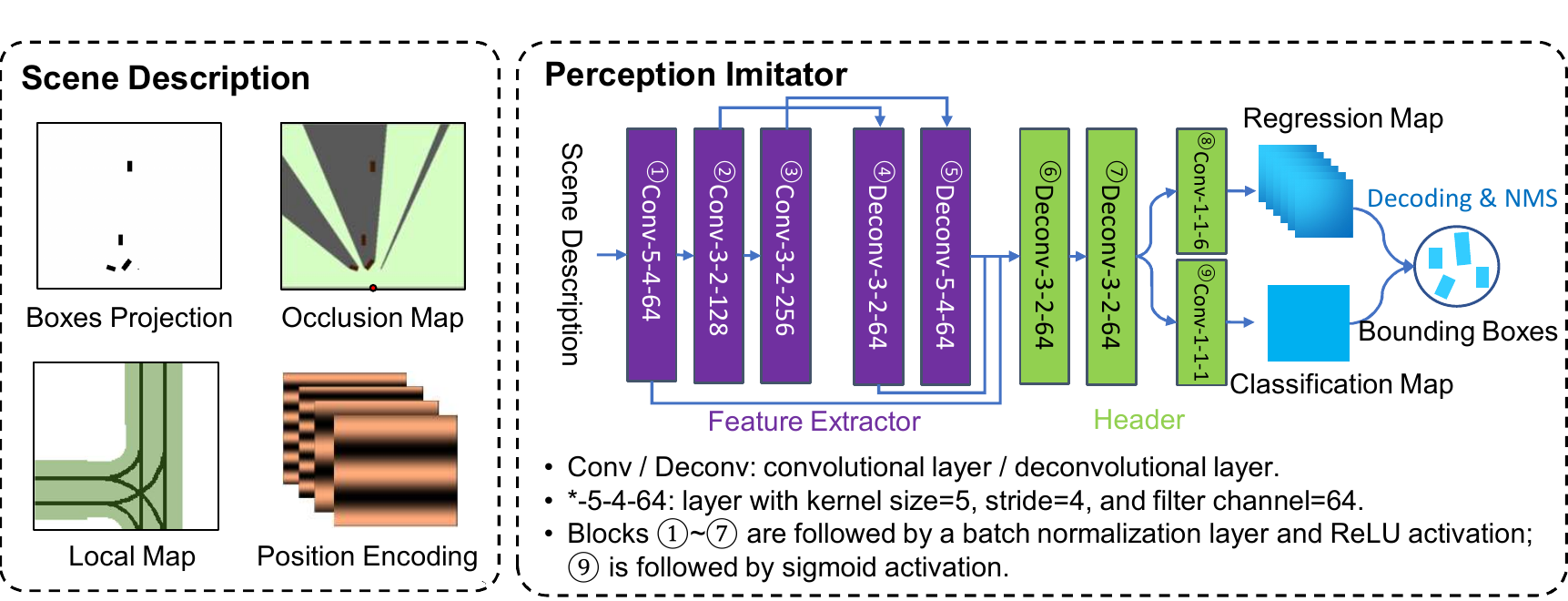}
    \caption{\red{Scene description and perception imitator.}
}
    \label{fig:network}
\end{figure}

As the previous formulation, perception imitator is the key component of our \red{synthesis-free} simulation.  It takes scene description as input, and outputs the bounding box coordinates as the simulated perception results.
\red{Compared to the previous work~\cite{wong2020testing}, our model optimizes the scene description by adding object position encoding, and exploits more efficient loss function to achieve a performance leap.}
In this section, we will expand on the design of scene description, the imitator model structure, and the learning process, as Figure \ref{fig:network} \& \ref{fig:model} show. 
\label{sec:method}

\subsection{Scene Description}
\label{subsec:scene description}
\red{A multi-channel map of road, vehicles, occlusion and positional encoding is used to represent the scene. Different from the baseline model~\cite{wong2020testing} depending on a multi-frame input, we only use a single frame scene description, and the positional encoding is added to involve the geometry feature.
}
\subsubsection{Local Road Map}
The map is rendered under ego coordinate as bird-eye-view image in 2 channels, including free space and the line formed by way-points. 
The position of ego vehicle in the map is the center of bottom margin.
All pixels take binary value, e.g. the background pixels equal to 0, and the free space pixels equal to 1.
\begin{figure*}[!h]
    \centering
    \includegraphics[trim={2.7cm 10.4cm 6.8cm 5.7cm},clip, width=17.5cm]{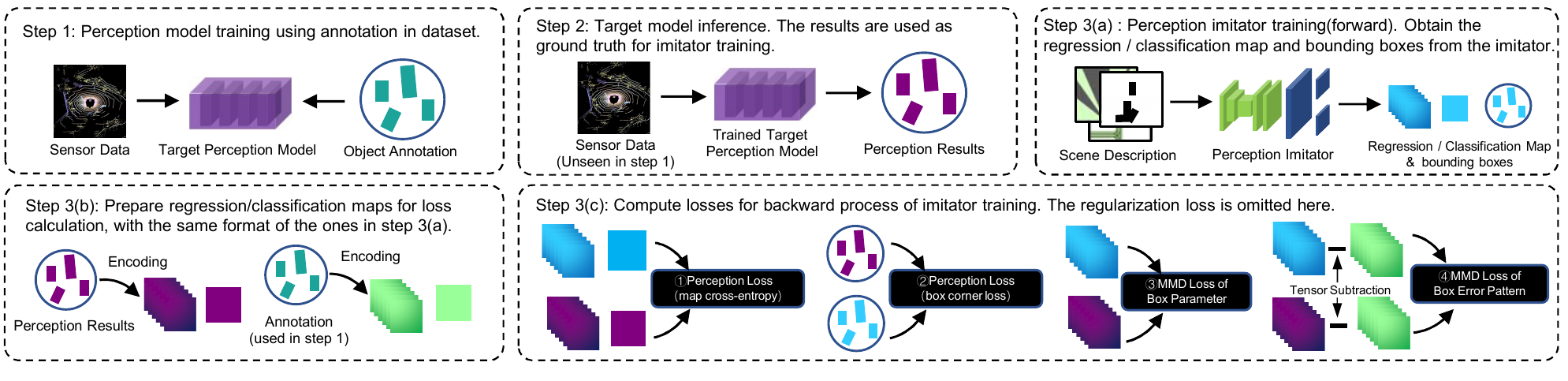}
    \caption{\red{Pipeline of perception imitator training. The perception results are used as ground truth to train perception imitator. To avoid confusion, the ground truth for target perception model training is called as annotation throughout the paper.}}
    \label{fig:model}
    \vspace{-4mm}
\end{figure*}

\subsubsection{Vehicles and Occlusion Map}
The vehicles are described as bounding boxes from a bird-eye-view and projected to an occupancy map, within the same coordinate of the local road map. \red{Intuitively, the mutual occlusion is an important factor to the perception performance. Therefore, to make it easier for the network to learn the occlusion feature, a binary occlusion map is built using a simple ray-casting algorithm, where every pixel represents the visibility from the view of LiDAR.}


\subsubsection{Positional Encoding}
The relative position of vehicles in the local scene is another important factor for perception. 
\red{To better explore such position clues, the position embedding is added as a supplementary channel. 
For faster convergence during model training, we only encode the image pixels along the heading direction of ego vehicle.}
Similar to \cite{vaswani2017attention}, we use sine and cosine functions of different frequencies.
\begin{equation}
\begin{aligned}
    PE_{pos, 2i}&=\sin(pos / 10000^{2i/d_{model}})\\
    PE_{pos, 2i+1}&=\cos(pos / 10000^{2i/d_{model}})
\end{aligned}
\end{equation}
where $pos$ is the relative longitude position to the ego vehicle and $i$ is the dimension, which means each dimension of the positional encoding
corresponds to a sinusoid. Here we choose the dimension size as $d_{model}=64$.

\red{Note that this simple set of scene description is a quick trial for our simulation pipeline. It works because it contains all the information about the objects that the perception model is detecting, such as the bounding boxes.  It can be understood as an error prediction of the target perception model. 
In fact, the factors affecting perception results are extremely complex and even uninterpretable. In real applications, it is necessary to optimize this description set continuously in a cycle of enumeration and evaluation.}


\subsection{Perception Imitation Model}
\label{method:pimodel}
Based on the image representation of the local scene, the task of perception imitator is to produce simulated perception results, subjected to similar distribution of a target model. 
It can be seen as a detection problem fed on scene sketches. Therefore, detection framework can be naturally exploited. \red{Here we use a one-stage model based on \cite{yang2018pixor}\cite{wong2020testing}, including a feature extractor and a header for object detection.}

\red{The network structure of our imitator is as Figure \ref{fig:network} shows. A cascade convolutional network with top-down branches is employed as our feature extractor, followed by a header with two branches for object classification and regression.
The classification branch is used to output a map of label confidence, and the regression branch serves to regress the parameters of the corresponding bounding box by producing a map of 6 channels including $(dx, dy, \log(w), \log(l), sin, cos)$, where $(dx, dy)$ denotes the pixel offset of the object center, $w, l$ denotes the object width and length, and $sin/cos$ denotes the $\tan(\theta)$.
As such, the bounding boxes can be decoded from the regression map, and then they can be filtered by confidence threshold and the non-maximum suppression. Please refer to \cite{yang2018pixor} for more details.}

\red{The network structure above has been widely used in the area of object detection in the previous years. We do not explore too much on the structure design, because we believe the capacity of such network is enough for learning from those simple scene descriptions, and the loss function design in the learning process maybe the more important aspect.}

\subsection{Learning}
\label{method:learning}
 \red{The learning steps of our perception imitator are illustrated in Figure \ref{fig:model}. First, a perception model is selected as perception target and trained on the desired data domain as the step 1. Second, its forward inference result is derived as the ground truth for imitator training as the step 2. Note that the target perception model may behave differently on the training data and the unseen data. It is more meaningful to imitate its behavior on the unseen part, so that only unseen data is used in this step.
 Then the imitator can be trained as steps 3(a)$\sim$(c) in Figure \ref{fig:model}. In the forward pass of training process as step 3(a), the intermediate classification map, regression map and the final bounding boxes are obtained. To prepare for the loss computation, the pseudo classification \& regression maps need to be derived from annotation and perception results as step 3(b), with same format and definition as the ones from our imitator. Then the loss items can be computed as step 3(c).
 }
 
 The loss function $L$ is composed of these parts: perception regression loss $L_p$, MMD~(maximum mean discrepancy) loss $L_m$ and regularization loss $\lambda\|w\|_2^2$.
\begin{equation}
    L=L_p + L_m + \lambda\|w\|_2^2
\end{equation}
The regularization loss $\lambda\|w\|_2^2$ is added to avoid over-fitting, especially on small dataset we used such as data collected from CARLA, where $\lambda$ is a hyperparameter set by experiment, $w$ is the parameter set of this model. The details of another 2 loss items  $L_p$ and $L_m$ are as follows.
\subsubsection{Perception Regression Loss}
$L_p$ is used to supervise the regression of all box instances.
\begin{equation}
    L_p = \alpha L_{cls} + \beta L_{reg} + \gamma L_{corner}
\end{equation}
where $L_{cls}$ is a binary cross entropy loss for classification, $L_{reg}$ is a smooth $L_1$ loss for box parameters regression,  $L_{corner}$ is an additional smooth $L_1$ item to limit the 4 corner positions of the boxes, and $\alpha, \beta, \gamma$ are hyperparameters for balance. Their definitions are listed below,
\begin{equation}
 L_{cls} = - \left\lbrack {Y_{cls}{\log X_{cls}} + \left( {1 - Y_{cls}} \right){\log\left( {1 - X_{cls}} \right)}} \right\rbrack 
\end{equation}
\begin{equation}
     L_{reg} = {smooth}_{L_{1}}\left\lbrack {Y_{cls}\left( {X_{reg} - Y_{reg}} \right)} \right\rbrack 
\end{equation}
\begin{equation}
     L_{corner} = \frac{1}{N}{\sum\limits_{i =1}^N{\sum\limits_{m = 1}^{4}\left\| {B_{i}^{m} - \hat{B}_{i}^{m}} \right\|}}
\end{equation}
where $ X_{cls}$ is the classification output from the object localization branch, $Y_{cls}$ is the corresponding ground truth, and $X_{reg}$ is the output from the regression branch with $Y_{reg}$ as its ground truth. Assume there are $N$ pairs of matched boxes, $\hat{B}_i^m$ is the $m$-th corner coordinate of the $i$-th bounding box decoded from $X_{i}$, with $B_i$ as its ground truth box matched by overlap. Here we want to remind the readers again that these ground truths are actually the prediction results from the target perception model.

\subsubsection{Maximum Mean Discrepancy Loss}
The objective of the MMD loss is to make the distribution of simulated perception results closer to the target one. MMD is a common measurement of two similar but different distributions, which is popularly used in generative models training to compare the generated samples and the target distribution. We use it as a loss item because it leads to more stable training process compared with the adversarial learning with a discriminator \cite{kar2019meta}. The equation of the MMD loss is
\begin{equation}
   L_{MMD^2}=\omega_1\sum_{l\in l_{reg}}\|M_{box}^l\|_\mathcal{H} + \omega_2\sum_{l\in l_{reg}}\|M_{err}^l\|_\mathcal{H}
\end{equation}
where
\begin{equation}
\begin{aligned}
M_{box}^l &= \frac{1}{n^2}\sum_{i,j}k(X_i^l, X_j^l) + \frac{1}{n^2}\sum_{i,j}k(Y_i^l, Y_j^l)\\
&-\frac{2}{n^2}\sum_{i,j}k(X_i^l, Y_j^l)\\
M_{err}^l &= \frac{1}{n^2}\sum_{i,j}k(X_i^l- Z_i^l, X_j^l- Z_j^l)\\
&-\frac{2}{n^2}\sum_{i,j}k(X_i^l-Z_i^l, Y_j^l-Z_j^l)\\
&+\frac{1}{n^2}\sum_{i,j}k(Y_i^l-Z_i^l, Y_j^l-Z_j^l)
\end{aligned}
\end{equation}

It consists of two parts balanced by the experimental hyperparameters $\omega_1, \omega_2$. The first part of $L_{MMD^2}$ measures the distribution difference between $\{X_l\},\{Y_l\}$,  which are the regression maps from our perception imitator and the target model.  However, we concern not only about the similarity between the results of the imitator and the target model, but also about the error pattern with respect to the annotation. Hence, the second part is added to measure the error from  $\{X_l\},\{Y_l\}$ to the regression maps $\{Z_l\}$ from annotation.
Here $X_i^l, Y_i^l, Z_i^l \in \mathbb{R}$ denote the $i$-th pixel value of the respective feature map $X_l, Y_l, Z_l$, $l_{reg}$ denotes the channel set of the feature map, $n$ denotes the pixel number in total, and $\mathcal H$ denotes the Hilbert space the kernel function $k$ projects to. In our work,  $k$ is a Gaussian kernel, so that $\mathcal H=\mathbb{R}$.

\section{Experiment}
We evaluate our method on both object detection dataset and down-streaming planning task. 
The datasets are used to evaluate the perception results on well-defined metrics as most detection tasks do. The down-streaming task is selected to compare the perception from an application perspective, which helps to prove the feasibility of replacing the original perception model with our perception imitator in continuous simulation loop. 
\label{sec:result}

\begin{table}[!h]
\resizebox{1.0\linewidth}{!}{%
\centering
\begin{threeparttable}[b]
\begin{tabular}{@{}lccc@{}}
\toprule[0.5mm]
 Target Model &CARLA & KITTI & nuScenes \\
\midrule
PointPillar\cite{lang2019pointpillars} & 96.4 / 96.7 / 86.1 & 87.2 / 91.9 / 78.4 & 63.7 / 97.1 / 44.9 \\
PVRCNN\cite{shi2020pv} & 99.2 / 95.6 / 98.4 & 87.9 / 97.1 / 44.9 & 70.4 / 92.8 / 70.5  \\
CenterPoint\cite{yin2021center} & 90.6 / 98.0 / 90.0 & 86.5 / 94.2 / 83.8 & 81.4 / 82.0 / 72.5  \\
\bottomrule[0.5mm]
\end{tabular}
\begin{tablenotes}
	\item *Every tuple denotes mAP/Precision/Recall in percentage.  
\end{tablenotes}
\end{threeparttable}
}
\caption{Target Perception Model Performance}
\label{tab:tp}
\vspace{-7mm}
\end{table}

\begin{table*}[!t]
\begin{center}
\begin{threeparttable}[b]
\begin{tabular}{p{0.8cm}p{0.1cm} p{0.74cm}p{0.74cm}p{0.74cm}p{0.74cm}p{0.0cm} p{0.74cm}p{0.74cm}p{0.74cm}p{0.74cm}p{0.0cm} p{0.74cm}p{0.74cm}p{0.74cm}p{0.74cm}}
\toprule[0.5mm]
\multirow{3}{*}{Dataset} & \multicolumn{1}{c}{\multirow{3}{*}{\begin{tabular}[c]{@{}l@{}}Perception\\Source\end{tabular}}}  
                         & \multicolumn{4}{c}{PointPillar} 
                         & \multicolumn{1}{c}{} 
                         & \multicolumn{4}{c}{PVRCNN}     
                         & \multicolumn{1}{c}{} 
                         & \multicolumn{4}{c}{CenterPoint} \\
                         & \multicolumn{1}{c}{} 
                         & \begin{tabular}[c]{@{}l@{}}(mA)P\%\\ (0.5)\end{tabular} 
                         & \begin{tabular}[c]{@{}l@{}}(mA)P\%\\ (0.7)\end{tabular} 
                         & \begin{tabular}[c]{@{}l@{}}(max)R\%\\ (0.5)\end{tabular} 
                         & \begin{tabular}[c]{@{}l@{}}(max)R\%\\ (0.7)\end{tabular}
                         & \multicolumn{1}{c}{} 
                         
                         & \begin{tabular}[c]{@{}l@{}}(mA)P\%\\ (0.5)\end{tabular} 
                         & \begin{tabular}[c]{@{}l@{}}(mA)P\%\\ (0.7)\end{tabular} 
                         & \begin{tabular}[c]{@{}l@{}}(max)R\%\\ (0.5)\end{tabular} 
                         & \begin{tabular}[c]{@{}l@{}}(max)R\%\\ (0.7)\end{tabular}
                         & \multicolumn{1}{c}{} 
                         
                         & \begin{tabular}[c]{@{}l@{}}(mA)P\%\\ (0.5)\end{tabular} 
                         & \begin{tabular}[c]{@{}l@{}}(mA)P\%\\ (0.7)\end{tabular} 
                         & \begin{tabular}[c]{@{}l@{}}(max)R\%\\ (0.5)\end{tabular} 
                         & \begin{tabular}[c]{@{}l@{}}(max)R\%\\ (0.7)\end{tabular} \\ \midrule
\multirow{5}{*}{CARLA}  
                         &  Gaussian                                         & 11.5 & 3.1 & 58.3 & 29.0     && 13.5 & 3.3 & 56.2 & 27.2     && 13.6 & 3.5 & 59.7 & 29.4 \\
                         &  Multimodal                                       & 15.6 & 7.4 & 68.4 & 46.3     && 19.3 & 8.4 & 64.5 & 44.1     && 16.2 & 7.6 & 66.3 & 44.9 \\
                         &  Baseline\cite{wong2020testing}                                         & 62.4 & 35.3 & 80.4 & 55.4    && 71.7 & 43.3 & 84.5 & 55.9    && 60.8 & 30.8 & 76.3 & 51.1 \\ 
                         &   Ours                                            & 89.3 & 68.1 & 94.1 & 80.4    && 93.9 & 69.7 & 95.1 & 79.1    && 91.8 & 73.1 & 96.0 & 84.7 \\ \cmidrule(l){2-16}
                          &  Annotation*                                               & 62.4 & 61.8 & 96.5 & 95.9    && 73.2 & 72.4 & 95.8 & 95.3    && 71.2 & 68.1 & 98.7 & 97.0 \\
                         \midrule
\multirow{5}{*}{KITTI}   
                         &  Gaussian                                         & 26.0 & 10.0 & 51.2 & 32.2    && 27.5 & 11.0 & 43.4 & 27.1    && 20.7 & 7.5 & 46.0 & 27.8 \\
                         &  Multimodal                                       & 34.3 & 27.1 & 60.0 & 53.4    && 35.6 & 26.6 & 49.2 & 42.5    && 42.2 & 29.8 & 65.8 & 55.1 \\
                         &  Baseline\cite{wong2020testing}                                         & 55.4 & 42.1 & 72.3 & 61.2    && 51.9 & 28.1 & 67.7 & 49.4    && 56.6 & 35.5 & 71.7 & 54.2 \\ 
                         & Ours                                              & 76.7 & 65.6 & 87.8 & 78.5    && 71.2 & 58.2 & 76.2 & 67.7    && 77.2 & 65.8 & 87.3 & 79.0 \\ \cmidrule(l){2-16}
                         &  Annotation                                               & 55.3 & 51.1 & 98.8 & 94.9    && 61.8 & 52.9 & 83.5 & 77.4    && 61.8 & 55.7 & 95.4 & 90.5 \\
                         \midrule
\multirow{5}{*}{nuScenes}
                         & Gaussian                                          & 3.1 & 0.20 & 40.0 & 10.2     && 1.6 & 0.30 & 36.4 & 16.8     && 2.1 & 0.30 & 20.3 & 7.5 \\
                         & Multimodal                                        & 5.2 & 1.8 & 51.9 & 29.9      && 3.8 & 0.60 & 55.4 & 21.8     && 4.1 & 0.40 & 28.2 & 9.2 \\
                         & Baseline\cite{wong2020testing}                                          & 51.0 & 38.4 & 66.4 & 55.2    && 48.1 & 31.6 & 63.6 & 49.1    && 46.8 & 26.6 & 66.4 & 48.7 \\ 
                         & Ours                                              & 64.7 & 51.7 & 72.3 & 60.5    && 54.5 & 39.3 & 57.9 & 47.3    && 58.8 & 41.2 & 63.3 & 49.8 \\  \cmidrule(l){2-16}
                         & Annotation                                                & 40.6 & 36.3 & 92.1 & 87.4    && 47.4 & 39.6 & 73.4 & 67.1    && 52.0 & 42.8 & 83.0 & 75.6 \\
                         
                         \bottomrule[0.5mm] 
\end{tabular}
\begin{tablenotes}
	\item *Annotation means the object box labels in datasets for training the target perception models.  Unlike the mAP/maxR values of Baseline/Ours group are calculated over recall values obtained from different score thresholds, the figures in Annotation/Gaussian/Multimodal group means only precision and recall because there is no score thresholding process involved. They are placed in same columns for convenience of comparison.

\end{tablenotes}
\end{threeparttable}

\end{center}

\caption{\red{Evaluation on Detection Metric}}
\label{tab:exp-det}
\vspace{-6mm}
\end{table*}

\subsection{Preparation}
\label{subsec:tpm}
\subsubsection{Dataset}
One synthesized dataset collected from CARLA~\cite{dosovitskiy2017carla} simulator and two public datasets KITTI~\cite{Geiger2012CVPR} and nuScenes~\cite{nuscenes} are used in our experiments.
We collected about 10k frames from map TOWN 01$\sim$05 for CARLA dataset, where vehicles and pedestrians are spawned randomly. 
These three datasets are with increasing perception difficulty, which helps to train  target perception models with different accuracy and validate the generalization ability of our imitator model.



\subsubsection{Target Perception Model}

With datasets ready, the next step is to train different target perception models on different datasets. \red{ 80\%, 50\% and 70\% of CARLA, KITTI and nuScenes are used in training, and the residuals are use for 1) testing the performance of those target models, 2) training and testing the perception imitator in subsection \ref{subsec:pit}}. The details of training setting are determined with the help of \cite{yin2021center}\cite{lang2019pointpillars}\cite{yin2021center}\cite{shi2020pv}\cite{openpcdet2020}. The performance described by mean-average-precision~(mAP)/precision/recall of target models are listed in Table \ref{tab:tp}, where mAP reflects the overall performance when different score thresholds adopted, and the following precision and recall corresponds to the score threshold we fixed for perception imitator.





\subsection{Perception Imitator Training}
\label{subsec:pit}
The inference results of the target perception models are dumped as the "ground truth" for perception imitator training.
\red{As mentioned before at the beginning of subsection \ref{method:learning}, it is more meaningful to imitate their behavior on unseen data, so that
we divide the unseen data in subsection \ref{subsec:tpm} into the new training/testing part for perception imitator by a proportion of 4:1.}
The image size in scene description is set to $352\times 400$, with pixel size $0.2{\rm m}\times 0.2{\rm m}$. Note that because KITTI dataset does not provide HDmap information, these map channels have been discarded in related experiments. The training time on three datasets are 300, 300, 500 epochs respectively, with Adam optimizer and a learning rate of 0.001.
The hyperparameters in subsection \ref{method:learning} are set by multiple experimental trials, and we finally fix $\lambda=0.001$, $\alpha=2.0$, $\beta=0.01$, $\gamma=10.0$, $\omega_1=0.005$, $\omega_2=0.001$.

\subsection{Results of Matching Quality}
\subsubsection{Metric Definition}
As we mentioned before, the perception imitator can be seen as a detector targeting on the inference results of other perception model, so that the frequently used metric for object detection is also reasonable for this special detector. Hence, we use mAP~(mean average precision) and maxR~(max recall) under different overlaps~(0.5 or 0.7) to measure the matching quality.

\subsubsection{Quantitative Analysis}

Five kinds of perception results are compared as the second column of Table \ref{tab:exp-det} shows. 
In order not to be confused with the ground truth we used in training perception imitator, we use {\it Annotation} to refer to the object labels in datasets. 
{\it Gaussian} means a simple noise model which adds Gaussian noise~($\mathcal{N}(0,0.1)$) to the results of regression branch of the imitator model. 
{\it Multimodal} denotes a noise model using mixture of Gaussian. We use {\it scikit-learn} toolbox for parameter fitting.
To simulate the FN~(false negative) cases, the boxes are randomly dropped according to FN ratio of target model in these two noise adding methods. 
\red{{\it Baseline} is as the description in subsection \ref{sub:pi}.}
More details about {\it Gaussian}, {\it Multimodal} and {\it Baseline} can be found in \cite{wong2020testing}.

As Table \ref{tab:exp-det} shows, the overall results corresponding to the {\it Gaussian} and {\it Multimodal}  are much worse than the other methods.
The reason is that these methods drop FN objects randomly, and the noise added to orientation may lead to failure of matching by overlap. 
\blue{It is worth noting that {\it Annotation} group is an important reference to judge whether an imitation method learns enough knowledge, because the {\it Annotation} information is actually embedded in the input of all these methods. In most cases, the performance of {\it Baseline} is not better than the {\it Annotation} group, while our perception imitator produces more satisfying prediction with an acceptable drop of recall.} Combining Table \ref{tab:tp} and Table \ref{tab:exp-det}, we can see that simpler dataset leads to better performance of perception model, and further results in better perception imitator. We attribute this phenomenon to the possibility that the perception model with better performance has a simpler error pattern, which may be related to hot topics of neural model uncertainty, so we will not discuss it further. 
\begin{figure*}[!t]
	\centering
	\includegraphics[trim={1.7cm 0.4cm 3.5cm 0.3cm},clip, width=17.5cm]{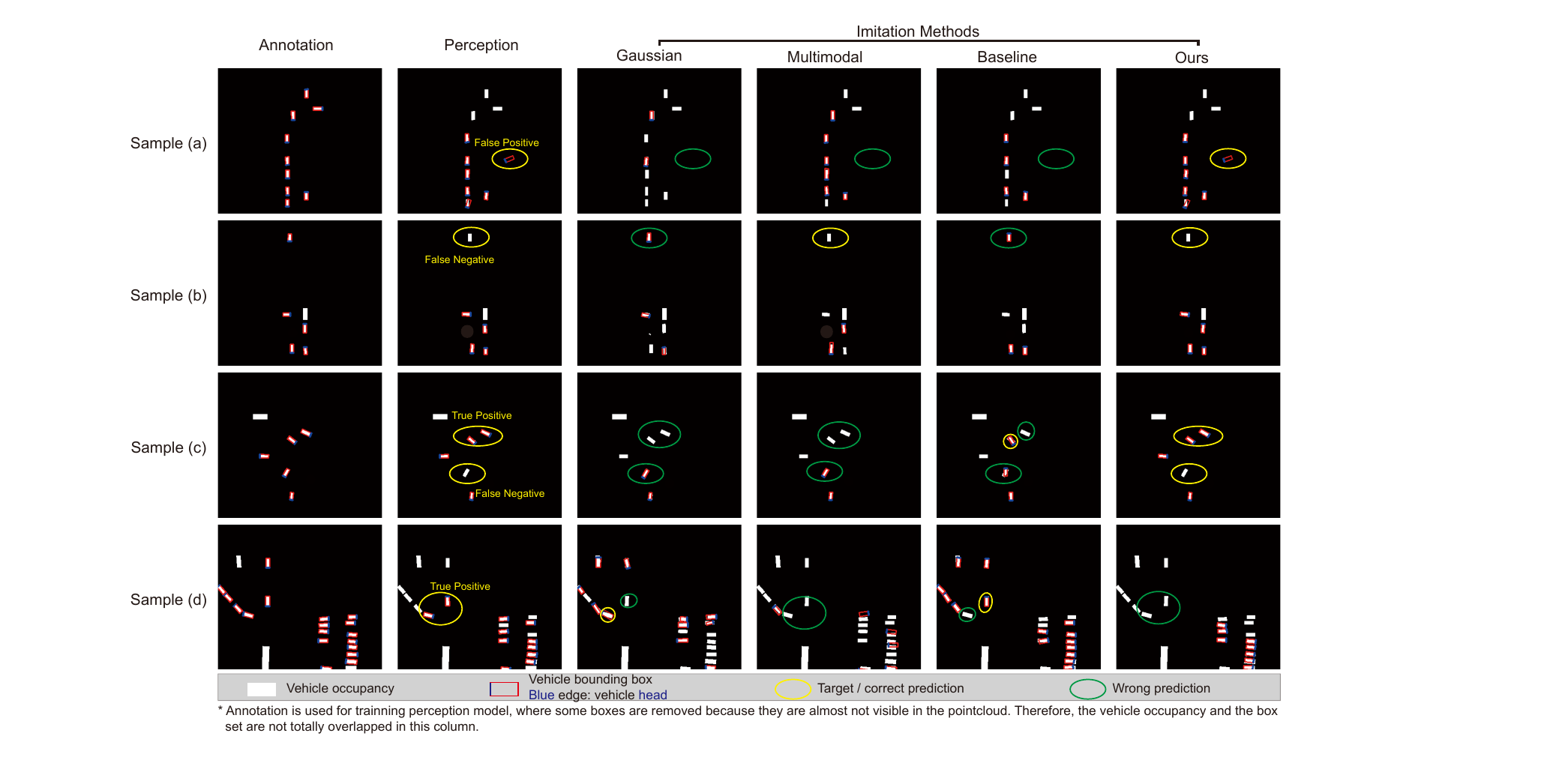}
	\caption{\red{Scene samples of comparison between different perception imitation methods. }}
	\label{fig:scene-examples}
	\vspace{-4mm}
\end{figure*}
\subsubsection{Qualitative Analysis}To examine the model performance more intuitively, perception results of 4 scene cases (a)(b)(c)(d) are visualized in Figure \ref{fig:scene-examples}. Case (a) is a representative that our model learns to produce similar false positive result as the target model, while (c) shows that our model can well identify which object to be judged as false negative. In addition, (a)-(d) indicate that our model learns to filter distant objects naturally, which also corresponds to a common phenomenon, that is, the more distant the object is, the more difficult it is to detect. However, these figures also give examples that our model misses some objects that the target model detects. This cannot be entirely attributed to the learning ability of the perception imitator model. For one thing, the input of the imitator is strictly limited to a simple enumeration of scene descriptions. For another, there exists a certain degree of randomness in the behavior of the target perception model, e.g. the perception models with same structure may behave slight differently when you trained an ensemble of them with similar settings.




\subsection{Evaluation by Downstream Task}
In this subsection, we want to examine that whether our perception imitator can be a substitute of the original perception model in downstream task, especially the interactive ones. 
This is because the proposed simulation route is where our perception imitator contributes. But for non-interactive tasks, the R\&D needs can be satisfied by datasets in most cases, so that simulator is not necessary. Therefore, we choose reinforcement learning based planning as our downstream task, and the evaluation method is as follows.

\subsubsection{Evaluation Scheme} 
The planner performance is used as the metric of evaluating different perception sources, based on the assumption that similar perception results lead to similar planning performances.
We need to train three agents A, B and C using the annotation, target perception model and our imitator respectively, where A represents the most popular training fashion without consideration of perception noise.
Next, A, B and C will be tested under the
results of target perception model, which is exactly the way planning module is deployed in.
Intuitively, the performance of B would be the best one, because it is trained and tested under the same data pattern.
In this experiment, we want to observe whether the performance of C is closer to B than A.

\begin{figure}[!h]
    \centering
    \includegraphics[trim={7.4cm 8.7cm 5.9cm 6cm},clip, width=8.5cm]{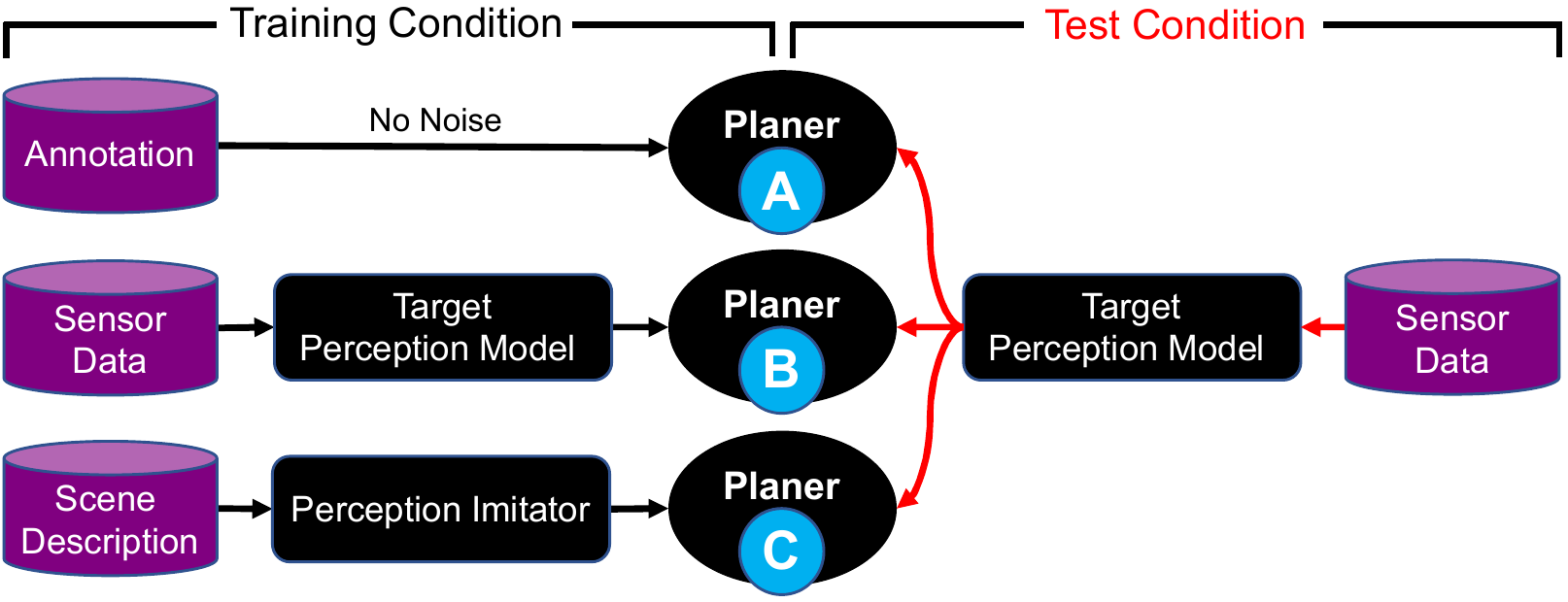}
    \caption{\red{Train planners as downstream task of perception imitation. }}
    \label{fig:ds_planning}
    \vspace{-2mm}
\end{figure}

However, to make the comparison above plausible, the sensor data of every simulation step is necessary for test of B. Therefore, we have to use synthesized data, which means the target model has to be trained on data produced by simulator. The whole evaluation scheme is shown in Figure \ref{fig:ds_planning}.

\subsubsection{Planning Task} 
CARLA is used as our simulation platform. The driving scene is simple, which is built on the outer cycle loop road of TOWN05. We expect the agent to have enough responses to the perception results, so that obstacles are placed as many as possible with the 30-meter minimum spacing. Inspired  by \cite{song2021autonomous}, we designed a simple MLP-based network to output action and value. In addition to the basic motion information~\cite{song2021autonomous} enumerates, the network input also includes the perception results converted into the range readings of different azimuths.
As for the reinforcement learning algorithm, PPO~\cite{schulman2015trust} is selected for its fast convergence. The whole system is established based on \cite{didrive}, which make it possible to run multiple CARLA server and collect data efficiently.
\begin{table}[!h]
\begin{tabular}{lcccc}
\toprule[0.5mm]
\multicolumn{1}{l}{Target Model / Planner} & \multicolumn{1}{c}{A} & B      & C(Ours) & C(Baseline) \\ \midrule
PointPillar & 209.09 & 266.82 & 237.04  & 104.32\\
PVRCNN & 211.97 & 290.02 & 246.55  & 123.58 \\
CenterPoint & 217.34 & 273.26 & 271.14 & 111.71\\
\bottomrule[0.5mm]
\end{tabular}
\caption{Test performance of planners: Average Rewards}
\label{tab:exp-planning}
\vspace{-7mm}
\end{table}
\subsubsection{Results}
Planners of types A, B and C are trained over 100K iterations, and the best checkpoints are selected for evaluation. The test scene is similar to the training setting, with obstacle vehicles placed randomly in every episode. For every type of agent, we average the rewards of 500 episodes as the planner performance, which actually equals to the distance it travels before collision. As Table \ref{tab:exp-planning} shows, the results are consistent with our expectations. It is not hard to explain that the planner group B outperforms group A, because planner group B is trained and tested under the same perception source. Planner group C using our perception imitator achieves closer performance than group A, which indicates the learned perception error pattern effects. We also do experiments on the planner group equipped with the baseline solution, whereas they may not be robust enough to reach good convergence.

\section{Conclusion}
In this work, we propose a novel \red{synthesis-free} simulation route for tasks independent of original sensor data.
The key component, a perception imitator model is designed to predict results of the target perception model under specific scene settings, and an abstract representation of the driving scene is devised as the input of our perception imitator. 

The proposed method is thoroughly evaluated on multiple datasets and target perception models in combination, with commonly used metric in object detection task. Also, a typical interactive downstream task is defined to test the perception imitator in continuous simulation loop. The experiments demonstrate that our perception imitator is able to reach approaching performance compared with original perception model, which naturally indicates the proposed simulation route is plausible and promising.

\bibliographystyle{IEEETran}
\bibliography{example.bib}  

\end{document}